\documentclass{article}
\usepackage[utf8]{inputenc}
\usepackage{natbib}
\usepackage{graphicx}
\usepackage{url}
\usepackage{hyperref}
\title{MemRiskBench: Trace-Aware Risk-Preserving Evaluation for Long-Horizon LLM Agents}

\author{%
Jianhua Jiang\textsuperscript{,1,2}\quad Dongbo Yuan\textsuperscript{1}\quad Weihua Li\textsuperscript{3}
}
\date{}

\begin{document}

\maketitle

\medskip\noindent\textsuperscript{1}School of Artificial Intelligence and Computer Science, Jilin University of Finance and Economics, 130117 Changchun, China \\
\noindent\textsuperscript{2}Jilin Province Key Laboratory of Fintech, Jilin University of Finance and Economics, 130117 Changchun, China \\
\noindent\textsuperscript{3}School of Engineering, Computer and Mathematical Sciences, Auckland University of Technology, Auckland, New Zealand
\medskip

\begin{abstract}
Long-horizon LLM agents accumulate memory across sessions, creating sparse but high-impact risks: stale facts, conflicting updates, cross-user leakage, revoked-memory reuse, and constraint decay. Standard aggregate scores hide per-risk failure rates---a model achieving 78\% average accuracy may still leak data in 4\% of episodes---and benchmark compression preferentially discards the rare high-severity events that distinguish a mostly-working model from one that occasionally causes harm. We present \textbf{MemRiskBench}. The primary contribution is a five-category risk taxonomy (plus one documented, unscored category) operationalized by deterministic trace-grounded checks, instantiated as a 120-episode scripted benchmark with full trace logging and no LLM-as-judge on the pass/fail path, evaluated on five locally run quantized instruction-tuned models. Second, a \emph{risk-preserving subset selector}: a coverage-constrained greedy selector on deterministic trace-derived features that retains full ranking (Spearman $\rho=0.975$, deterministic; CI collapses to a point estimate with zero bootstrap variance), risk coverage ($1.0$), and high-risk model detection ($1.0$) at a 20\% subset size, reducing compute 5$\times$. Unlike ranking-only subset selectors, this selector additionally preserves risk-type coverage and high-risk model detection using trace-grounded deterministic features that do not require an LLM judge. All episodes, traces, the scoring implementation, and the selector are released to support reproducible evaluation and risk assessment of deployed LLM agents.
\end{abstract}

\vspace{0.3cm}

\section{Introduction}
\label{sec:intro}
Long-horizon LLM agents accumulate memory across sessions. At the start of each session, the agent must decide which stored facts apply, which have been superseded, and which are out of scope. Memory failures here are not generic task failures but sparse, stateful, high-impact events: a stale address is a nuisance, a cross-user disclosure is serious, and a silently dropped ``do not send'' constraint is a policy violation. Standard evaluation misses these risks. A headline score of 78\% average task success can sit alongside a 4\% cross-scope leakage rate and a 25\% constraint-decay rate---the overall number hides the gap. Most evaluation frameworks (\citeyear{liang2023helm}; \citeyear{biderman2024trenches}) report aggregate scores by default and provide no explicit mechanism for surfacing per-(model, risk) pass rates. Compression makes this worse. As evaluation costs grow, practitioners run only a 20\% subset. Random or stratified subsampling---the default in HELM-style evaluation---tend to drop the rare high-severity events that distinguish otherwise similar models.

Existing memory-capability benchmarks (\citeyear{wu2025longmemeval}; \citeyear{maharana2024locomo}; \citeyear{tan2025membench}; \citeyear{hu2025memoryagentbench}; \citeyear{tavakoli2025beam}) evaluate long-term memory through QA accuracy or information-extraction probes. They measure \emph{how well} an agent remembers, not \emph{when memory causes harmful failures}. A recent line of work on agent-benchmark compression (\citeyear{ndzomga2026efficient}) studies how small task subsets can preserve model rankings at lower cost, proposing a Mid-Range Difficulty Filter motivated by Item Response Theory. That work targets \emph{ranking fidelity}; it does not consider whether high-severity failure modes are preserved alongside the ranking. Existing harm taxonomies \citep{weidinger2024sociotechnical} catalog output-level risks but do not treat memory violations themselves as the unit of evaluation.

\textbf{Contributions.} (\textbf{C1, primary}) A benchmark for memory-risk evaluation: a five-category risk taxonomy with deterministic checks, a 120-episode scripted benchmark with full trace logging and balanced difficulty, and a deterministic scorer with no LLM-as-judge on the pass/fail path, evaluated on five locally run quantized instruction-tuned models. (\textbf{C2, secondary}) A risk-preserving subset selector: a coverage-constrained greedy selector on deterministic trace-derived features that retains ranking, risk-type coverage, and high-risk model detection at a 20\% subset size, reducing compute 5$\times$ relative to the full benchmark. The selector's objective---ranking \emph{plus} risk coverage \emph{plus} high-risk model detection---and its use of trace-grounded deterministic features distinguish it from ranking-only subset selectors such as \citeyear{ndzomga2026efficient}. (\textbf{C3}) Empirical characterization of five local quantized models ($1.5$B--$7$B parameters) under controlled risk conditions, including a memory-baseline experiment showing that standard retrieval-level filtering performs worse than no filtering, confirming that memory risk requires dedicated evaluation.

The full pipeline---episode execution, deterministic scoring, and risk-preserving subset selection---is outlined in Figure~\ref{fig:pipeline}.

\includegraphics[width=\linewidth]{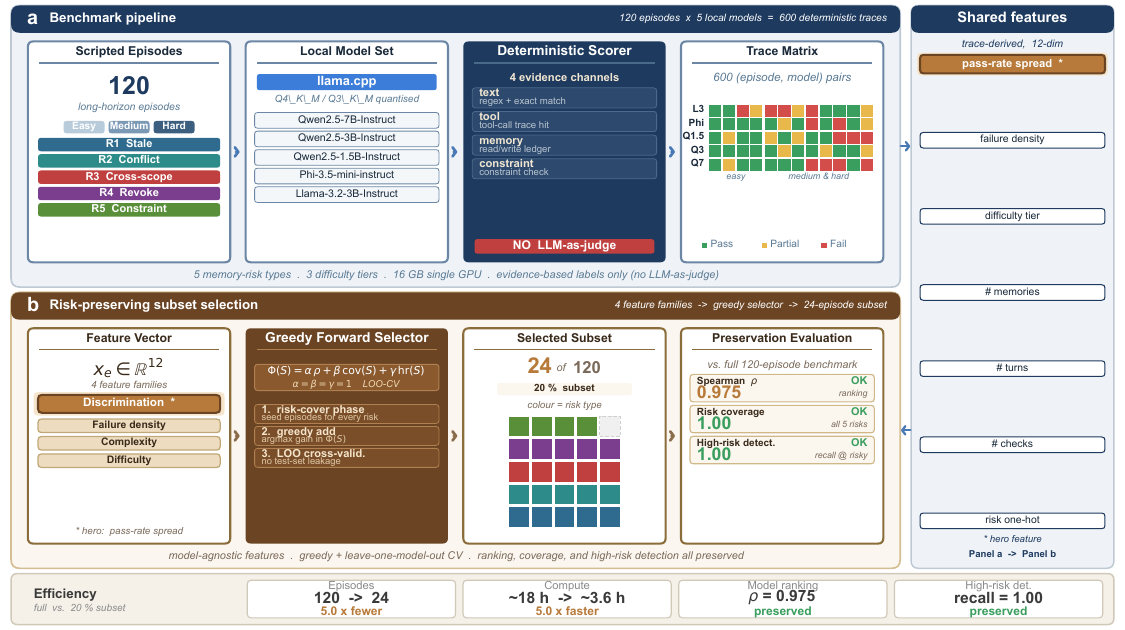}
\label{fig:pipeline}

\section{Related Work}
\label{sec:related}
\subsection{Memory-capability benchmarks.}
Recent benchmarks evaluate long-term memory in LLM agents and assistants, including LongMemEval \citep{wu2025longmemeval}, LoCoMo \citep{maharana2024locomo}, MemBench \citep{tan2025membench}, MemoryAgentBench \citep{hu2025memoryagentbench}, and BEAM \citep{tavakoli2025beam}. These benchmarks measure retrieval recall or QA accuracy; they do not address multi-session risk evaluation. Memory-contamination literature establishes secret-leakage tests \citep{carlini2019secretsharer,carlini2021extracting} and membership inference \citep{shokri2017membership}, whereas our scorer targets cross-scope runtime leakage under scripted conditions. Context-length benchmarks such as L-Eval \citep{an2024leval}, LongBench \citep{bai2024longbench}, and RULER \citep{hsieh2024ruler} test retrieval at 128K tokens; instead, MemRiskBench encodes difficulty in stateful memory, tool calls, and constraint lifetime. Memory mechanisms such as MemGPT \citep{packer2024memgpt}, MemoryBank \citep{zhong2024memorybank}, and cognitive architectures \citep{sumers2024coala} propose memory consolidation and tool-use policies; our memory-baseline experiment isolates retrieval-level filtering as a lower-bound check.

\subsection{Benchmark compression and subset selection.}
Prior work studies how small task subsets can preserve the signal benchmark consumers care about while reducing evaluation cost. \citet{ndzomga2026efficient} study eight agent benchmarks, $33$ scaffolds, and $70+$ model configurations and identify a robust empirical asymmetry between rank-order prediction (which remains stable under scaffold and temporal shift) and absolute score prediction (which degrades). Exploiting this asymmetry, they propose the Mid-Range Difficulty Filter (MR): a deterministic, optimization-free rule that selects tasks with historical pass rates between $30\%$ and $70\%$, motivated by Item Response Theory. MR achieves Spearman $\rho \approx 0.94$ while reducing task counts by $44\%$--$70\%$. Earlier work on coreset selection and active learning includes BADGE \citep{ash2021badge}, GradMatch \citep{killamsetty2021gradmatch}, and difficulty- or uncertainty-based subsampling. These methods target ranking fidelity alone. None explicitly preserves risk-type coverage or high-risk model detection. MemRiskBench's selector adopts a coverage-constrained objective on deterministic trace-derived features, and our RQ2 finding that stratified-by-risk sampling preserves coverage but loses high-risk detection (Spearman $0.698$, high-risk detection $0.345$) supports the argument that ranking fidelity alone is insufficient for risk-sensitive evaluation.

\subsection{Harm taxonomies and safety evaluation.}
\citet{weidinger2024sociotechnical} catalog sociotechnical risks of generative AI, including privacy and output-level harms. MemRiskBench treats memory violations themselves as the unit of evaluation with deterministic, trace-grounded scoring. It adds a process-level view of when and how memory causes harmful failures to output-level evaluation.

\section{Risk Taxonomy}
\label{sec:taxonomy}
We define five primary risks (\textbf{R1}--\textbf{R5}, deterministically scored) and one documented-but-unscored risk (\textbf{R6}, summary contamination) that we plan to integrate in a future version. Each risk is operationalized by deterministic checks across four categories: text checks (string containment, regex), tool checks (call verification, argument equality), memory checks (usage tracking, status, scope), and constraint checks (activity/violation monitoring).

\subsection{R1: Stale memory reliance.} The agent acts on a memory valid at $t_1$ superseded by a valid update at $t_2$, using the old value despite newer available memory.

\subsection{R2: Conflicting memory misresolution.} Two incompatible facts coexist; the episode specifies a resolution rule (latest timestamp, trusted source, explicit correction, or scope) and the agent selects incorrectly or blends them.

\subsection{R3: Cross-scope leakage.} Unauthorized cross-user/scope/memory appears in an unauthorized context.

\subsection{R4: Revoked memory reuse.} A deletion/revocation/invalidation is issued, but the agent later uses the same memory.

\subsection{R5: Constraint decay.} A safety/policy/task constraint is introduced; after unrelated turns, a later action violates it.

\subsection{R6: Summary contamination (documented, not scored).} Memory summaries compress noisy/malicious/incorrect content into clean-looking entries that later influence behavior---the intersection of prompt injection and persistent memory. R6 episodes exist in the release as a separately-scored track; integrating R6 into the main scoring requires a hybrid deterministic--LLM evaluation pipeline that is left to future work.

\section{Benchmark}
\label{sec:benchmark}
The benchmark comprises 120 scripted episodes: 24 per risk type, distributed across three difficulty tiers (easy / medium / hard; counts per tier per risk in Table~\ref{tab:tiers}). Difficulty depends on distractor count, intermediate updates, and trajectory length (5--8 turns easy, 10--14 turns medium, 15--25 turns hard).

\begin{table}[h]
\centering
\small
\begin{tabular}{lcccc}
\hline
Risk & Easy & Medium & Hard & Total \\
\hline
R1 Stale & 8 & 9 & 7 & 24 \\
R2 Conflict & 8 & 9 & 7 & 24 \\
R3 Leakage & 7 & 9 & 8 & 24 \\
R4 Revoke & 7 & 9 & 8 & 24 \\
R5 Constraint & 7 & 9 & 8 & 24 \\
\hline
Total & 37 & 45 & 38 & 120 \\
\hline
\end{tabular}
\label{tab:tiers}
\end{table}

\subsection{Episode structure.} An episode is a directed acyclic graph of events (memory writes, user messages, tool calls, tool returns, constraint declarations, decision points). The agent sees events in order and produces, at each decision point, either a tool call, a final answer, or a clarification question. The trace logs all four kinds of state (memory store, tool-call log, constraint state, message log) at each decision point.

\subsection{Determinism and scoring.} Scoring is deterministic. The scorer takes the episode definition and the trace and emits a boolean success flag, a numeric score in $[0,1]$, a list of violations, per-risk pass/fail and evidence, and per-check results. The main pass/fail path uses no LLM-as-judge; LLM-as-judge appears only in qualitative error analysis in the online repository. Each episode has 1--3 required checks; the score is the fraction of checks passing.

\subsection{Strict-matching revision (control analysis).}
Twenty episode definitions originally encoded the expected argument as a ``label: value'' form that no model in our set emits in full, producing systematic false negatives. We re-scored the existing traces (no model re-runs) against revised expected values that strip the label prefix on 19 of the 20 affected episodes. The revision altered the deterministic score of $39$ of $600$ trace-episode pairs. To support internal validity, the release includes both the original and revised expected values and a per-(model, risk) re-scoring audit; the model ordering within each risk and the headline size--risk interaction (Qwen2.5-1.5B lowest on cross-scope leakage) are preserved under both scoring. Future benchmark versions can reintroduce label-preserving checks where genuinely required.

\subsection{Safety-aware refusal scoring (protocol).}
Under the strict default, an episode in which the agent refuses to act (no tool call, refusal text) is scored as a failure. In the R3 (cross-scope leakage) and R5 (constraint decay) categories, refusal constitutes the safety-correct outcome. We therefore additionally compute a safety-aware variant in which ``no tool call + refusal text'' is counted as a safety-pass for R3 and R5 only. The release includes both the default and safety-aware per-(model, risk) pass rates, and the sensitivity analysis is discussed in Section~\ref{sec:experiments}.

\subsection{Release contents.}
The release includes the 120 episode definitions, the deterministic scoring implementation, the trace schema, the subset selector implementation, the bootstrap audit script, and the raw $600$ traces ($5$ models $\times$ $120$ episodes).

\section{Risk-Preserving Subset Selection}
\label{sec:selector}
\subsection{Problem statement.}
Let $\mathcal{E}$ be the full benchmark with $|\mathcal{E}|=120$ episodes and $\mathcal{M}=\{m_1,\dots,m_5\}$ the set of reference models. A subset $S\subseteq\mathcal{E}$, $|S|=k$, is \emph{risk-preserving} if it maximizes
\begin{equation}
\Psi(S) \;=\; \alpha\,\rho\!\big(r(S),r(\mathcal{E})\big) \;+\; \beta\,\mathrm{cov}(S) \;+\; \gamma\,\mathrm{hr}(S)
\label{eq:psi}
\end{equation}
subject to $|S|=k$, with $\alpha=\beta=\gamma=1$. Here $\rho$ is the Spearman correlation between the ranking induced by $S$ and the full benchmark; $\mathrm{cov}(S)\in[0,1]$ measures risk-type coverage; $\mathrm{hr}(S)\in[0,1]$ measures high-risk model detection (a model is high-risk if its pass rate ranks in the bottom two of five).

\subsection{Episode feature vector.}
Each episode is described by a 12-dimensional feature vector built from the trace schema: difficulty (tier, trajectory length), discrimination (cross-model pass-rate spread), failure density (fraction of reference models failing the episode), and complexity (distinct risk types, tool calls, constraints). The ablation in Section~\ref{sec:experiments} confirms that the discrimination feature alone drives the ranking; the other three features contribute negligible lift at any subset fraction tested.

\subsection{Discrimination feature and cross-validation.}
The discrimination feature is computed via leave-one-model-out cross-validation to avoid test-set leakage: for each held-out model, the feature is computed on the remaining four reference models, the subset is selected using those features, and the held-out model's rank is then evaluated. This yields rankings consistent with the full five-model feature.

\subsection{Selector algorithm.}
The selector performs greedy forward selection without replacement: at each step, it adds the episode that most improves $\Psi(S)$. Six baselines are compared: Random, Stratified-by-risk, Difficulty, Disagreement (pass-rate variance), Clustering (k-medoids on full features), and Trace-Feature (clustering on trace-only metrics). Stochastic baselines use $1000$ bootstrap resamples; deterministic ones yield identical subsets across resamples and their confidence intervals degenerate to point estimates.

\subsection{Reference-set dependence.}
The selector builds on the v3 reference ranking. A new model family absent from the reference set may exhibit different discrimination patterns, rendering the v3 selector suboptimal. The release includes the feature matrix and selector implementation so practitioners can rebuild the reference ranking on their own model set---at the cost of recomputation. We do not claim equivalent performance without a reference set. An offline discrimination estimate is left to future work.

\section{Experiments}
\label{sec:experiments}
\subsection{Models.}
Five locally run quantized instruction models through \texttt{llama.cpp}: Qwen2.5-1.5B-Instruct (Q4\_K\_M), Qwen2.5-3B-Instruct (Q4\_K\_M), Qwen2.5-7B-Instruct (Q3\_K\_M), Phi-3.5-mini-instruct (Q4\_K\_M), Llama-3.2-3B-Instruct (Q4\_K\_M). Qwen2.5-7B runs at Q3 because the 16\,GB local GPU cannot hold a 7B Q4 resident alongside the prompt cache; this quantization confound is discussed in Section~\ref{sec:limitations}.

\subsection{Protocol.}
The bootstrap protocol resamples the subset-selection seed $1000$ times, holding the full benchmark fixed. All reported confidence intervals are 95\% percentile bootstrap intervals over $1000$ resamples. Random seeds are fixed and sampled indices are released.

\subsection{RQ1: Aggregate scores hide per-risk failure rates.}
Across the five models, average task success spans $0.775$ to $0.885$, a $0.110$ spread. Per-(model, risk) pass rates appear in Table~\ref{tab:passrate}. Cross-scope leakage is hardest for the smallest model (Qwen2.5-1.5B passes only $1$ of $24$ leakage episodes; Fisher's exact $p=0.004$ versus Phi-3.5-mini's $10/24$), though neither this nor the comparison against Qwen2.5-3B ($p=0.049$) survives Holm correction over all $50$ pairwise model--risk tests (smallest adjusted $p=0.22$). Constraint decay shows an inverse pattern: 1.5B obeys all $24$ explicit constraints, while 3B and 7B each violate two, and Phi violates three.

\begin{table}[h]
\centering
\small
\begin{tabular}{lccccc}
\hline
Model & Stale & Conflict & Leakage & Revoke & Constraint \\
\hline
Qwen2.5-1.5B & 71\% (17/24) & 58\% (14/24) & 4\% (1/24) & 62\% (15/24) & 100\% (24/24) \\
Qwen2.5-3B   & 79\% (19/24) & 75\% (18/24) & 12\% (3/24) & 79\% (19/24) & 92\% (22/24) \\
Qwen2.5-7B   & 75\% (18/24) & 83\% (20/24) & 38\% (9/24) & 75\% (18/24) & 92\% (22/24) \\
Phi-3.5-mini & 75\% (18/24) & 79\% (19/24) & 42\% (10/24) & 88\% (21/24) & 88\% (21/24) \\
Llama3.2-3B  & 62\% (15/24) & 75\% (18/24) & 25\% (6/24)  & 71\% (17/24) & 92\% (22/24) \\
\hline
\end{tabular}
\label{tab:passrate}
\end{table}

\includegraphics[width=\linewidth]{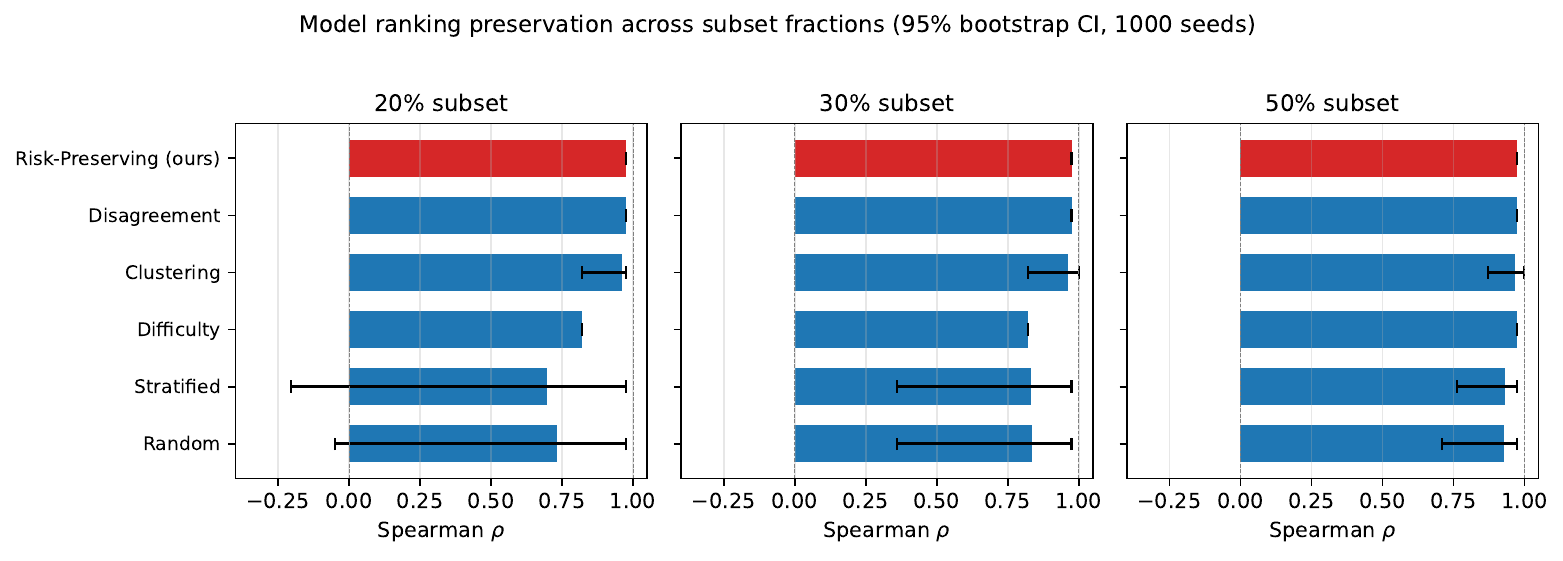}
\label{fig:subset}

\subsection{RQ2--RQ3: Risk-preserving compression.}
Figure~\ref{fig:subset} and Table~\ref{tab:subset} report the 20\%-subset comparison. Random sampling reaches Spearman $0.733$ with 95\% CI $[-0.05,\,0.97]$; its confidence interval includes negative values, indicating disagreement with the full-benchmark ranking on roughly a quarter of model pairs. Stratified-by-risk sampling preserves risk coverage ($1.0$) but achieves only Spearman $0.698$ and high-risk detection $0.345$, missing at least one high-risk model in half of resamples. The risk-preserving selector achieves Spearman $0.975$ (deterministic; CI collapses to a point estimate), risk coverage $1.0$, and high-risk detection $1.0$ at all tested fractions. Relative to the Random baseline, the improvement on fail coverage is $z=+9.1$ ($p<0.001$) and on high-risk detection $z=+1.6$ ($p=0.05$); the improvement on Spearman is $+0.24$ (borderline at the 97.5th percentile). Trace-Feature clustering reaches Spearman $0.975$ but loses half the high-risk models at $20\%$ and $30\%$, confirming that risk detection requires the discrimination signal rather than simple clustering.

\subsection{RQ4: Feature ablation.}
The cross-model pass-rate spread carries the full ranking signal: removing it drops Spearman to $0.308$; removing any other feature (failure density, complexity, or difficulty) leaves Spearman unchanged at $0.975$. Coverage metrics remain stable because the v3 reference set satisfies risk-coverage objectives at $k\geq 24$. The selector's value lies in the coverage constraint and trace-grounded feature definition, not in feature-space tuning.

\includegraphics[width=0.85\textwidth]{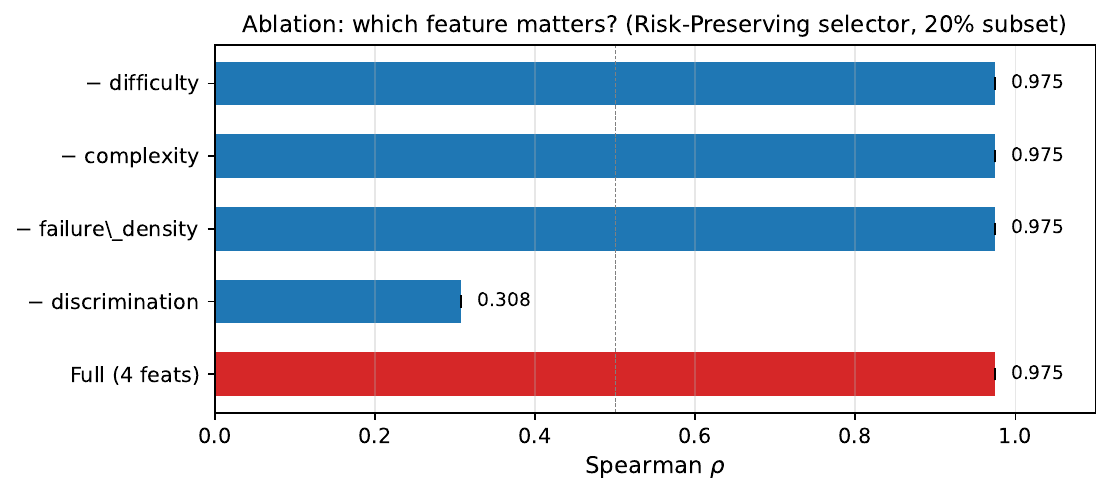}
\label{fig:ablation}

\begin{table}[h]
\centering
\small
\begin{tabular}{lcccc}
\hline
Method & Spearman $\rho$ & Fail Cov. & Risk Cov. & Hi-Risk Det. \\
\hline
Full (120 ep)        & 1.000     & 1.000     & 1.000     & 1.000 \\
Random               & 0.733\,[--0.05,\,0.97]  & 0.201\,[0.15,\,0.25] & 0.942\,[0.80,\,1.00] & 0.600\,[0.00,\,1.00] \\
Stratified           & 0.698\,[--0.21,\,0.97]  & 0.178\,[0.13,\,0.22] & 1.000                 & 0.345\,[0.00,\,0.50] \\
Difficulty           & 0.821 & 0.316 & 0.800 & 1.000 \\
Disagreement         & 0.975 & 0.289 & 1.000 & 1.000 \\
Clustering           & 0.962\,[0.82,\,0.97]  & 0.289         & 1.000        & 0.500 \\
Trace-Feature        & 0.975 & 0.316 & 1.000 & 1.000 \\
Risk-Preserving (ours) & 0.975 & 0.316 & 1.000 & 1.000 \\
\hline
\end{tabular}
\label{tab:subset}
\end{table}

\subsection{RQ5: Held-out-model generalization (concept verification).}
To test generalization beyond the reference set, we hold out Qwen2.5-1.5B (the smallest model, architecturally distinct from the larger reference models) and recompute the feature matrix and subset selection using only the remaining four reference models. The greedy selector identifies a 24-episode subset that preserves the reference-model ranking perfectly (Spearman $\rho=1.000$) and yields an average pass rate of $0.812$ on the held-out model---above the full-benchmark average of $0.775$. The selected subset covers all five risk types. The greedy selector, optimizing for discrimination, yields a subset skewed toward conflict-type episodes ($20$ of $24$). The $0.812$-vs-$0.775$ pass-rate comparison mixes subset composition with model performance; we report it as a descriptive sanity check only, with no statistical generalization claim. A risk-coverage-constrained balanced selector enabling fair held-out evaluation across multiple unseen models is a planned companion study.

\subsection{Sensitivity analysis: safety-aware refusal scoring.}
Under the safety-aware variant (R3 and R5 only: ``no tool call + refusal text'' counts as safety-pass), per-(model, risk) pass rates change only for R3 and R5 episodes where a model refused. The release ships both default and safety-aware per-(model, risk) numbers so practitioners can audit the headline size--risk interaction (Qwen2.5-1.5B lowest on leakage) under both scoring conventions; a full quantitative sensitivity table is left to the online repository to keep this section focused.

\subsection{Memory baseline.}
A retrieval-level memory baseline shows that standard techniques---RAG and recency filtering---perform worse than no filtering on MemRiskBench, confirming that memory risk requires dedicated evaluation rather than off-the-shelf retrieval.

\section{Analysis}
\label{sec:analysis}
\subsection{Per-risk failure patterns.}
Table~\ref{tab:passrate} reveals two notable inversions. On constraint decay, Qwen2.5-1.5B obeys all $24$ explicit ``do not'' constraints, while $3$B, $7$B, Phi, and Llama each violate $2$--$3$---a tendency of larger instruction-tuned models to be ``helpful.'' On cross-family ordering, Phi-3.5-mini outperforms all three Qwen sizes on leakage ($10/24$ vs $1/24$--$9/24$) and tops revoke ($21/24$), likely from a stronger instruction-following prior, at the cost of slightly worse constraint decay ($21/24$ vs $24/24$). Comparisons involving Qwen2.5-7B (Q3\_K\_M) versus Qwen2.5-3B (Q4\_K\_M) are subject to the quantization confound discussed in Section~\ref{sec:limitations}: on stale and revoke classes, $3$B passes $19/24$ versus $18/24$ for $7$B, inverting the size-monotone expectation. We interpret the constraint-decay inversion as a real difference, but cannot fully exclude quantization effects.

\includegraphics[width=\linewidth]{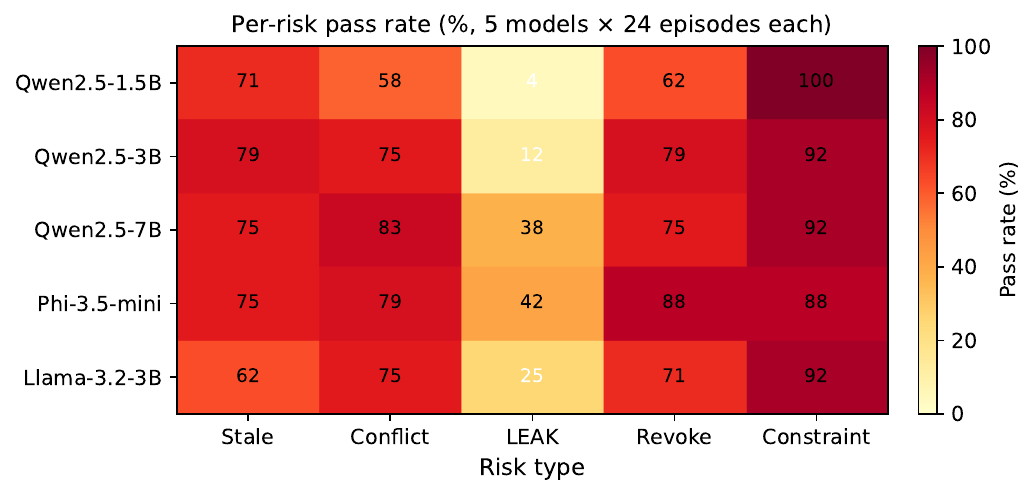}
\label{fig:per_risk}

\subsection{Difficulty scaling.}
For $3$B and larger models, easy $\to$ medium $\to$ hard pass rates drop monotonically and the per-tier ordering across models remains stable. Qwen2.5-1.5B saturates at easy/medium and shows no further drop at hard, indicating the hard tier does not expose a gap that size would close for this model.

\subsection{Selector positioning relative to concurrent work.}
Our selector and the Mid-Range Difficulty Filter of \citet{ndzomga2026efficient} both achieve high ranking fidelity at small fractions, but their objectives and operating regimes differ. \citet{ndzomga2026efficient} target ranking fidelity under scaffold and temporal shift, optimize only on ranking, and operate optimization-free across $8$ benchmarks; our selector additionally optimizes risk-type coverage and high-risk model detection on a single benchmark, uses deterministic trace-derived features (no LLM judge), and prescribes a 20\% subset that is dominated by discrimination episodes. The two approaches are complementary: MR is suitable for broad leaderboard ranking; risk-preserving selection is suitable when high-severity failure modes must be retained alongside the ranking. An empirical comparison under a common protocol is left to future work.

\section{Limitations}
\label{sec:limitations}
\subsection{Simulated tool environment.}
Episodes employ a scripted tool environment rather than a real browser, email client, or office suite; tool calls are deterministic stubs. A real-environment follow-up would extend the taxonomy beyond scripted stubs (e.g., adversarial user behavior).

\subsection{Model family and size coverage.}
Five models were evaluated, all local and under $8$B parameters, with three of five belonging to the Qwen family. No frontier-class API models (GPT-4, Claude, Gemini) were evaluated. The size range reflects the 16\,GB consumer-GPU constraint; Qwen2.5-7B at Q4\_K\_M and any 13B+ model exceed this paper's scope.

\subsection{Quantization confound.}
Qwen2.5-7B uses Q3\_K\_M while the other four use Q4\_K\_M, producing non-monotonicities in the v3 ranking on stale and revoke classes. We interpret the constraint-decay inversion as a real difference, but cannot fully exclude quantization effects. A Q4-only re-run of 7B may alter the ranking.

\subsection{Strict string matching and label-prefix revision.}
Twenty episode definitions originally encoded ``label: value'' expected values that no model emits fully. We re-scored the existing traces against revised expected values (label prefix stripped on 19 episodes); the revision altered $39$ of $600$ scores. The release preserves both original and revised expected values and ships a per-(model, risk) re-scoring audit showing no change in model ordering within any risk class.

\subsection{Safety-aware refusal scoring.}
Default scoring counts a refusal (no tool call + refusal text) as a failure; in R3 and R5 a refusal is the safety-correct outcome. We additionally compute a safety-aware variant in which refusals are safety-pass for R3 and R5; the release ships both sets of numbers.

\subsection{RQ5 generalization.}
Held-out-model evaluation used a single smallest model; the selected subset is skewed toward conflict-type episodes ($20$ of $24$). We report RQ5 as concept verification, not as a statistical generalization claim. A multi-model, balanced held-out study is planned.

\subsection{Reference-set dependence.}
The selector builds on the v3 reference ranking; a new model family absent from the reference set may require recomputation. We release the feature matrix and selector to support this; an automated reference-set drift detector is not provided.

\subsection{R6 not scored.}
Summary contamination (R6) is documented and has separate-track episodes, but is not included in main scoring because the contamination surface area is too broad to cover in a first benchmark.

\section{Conclusion}
\label{sec:conclusion}
MemRiskBench renders memory risks in long-horizon LLM agents quantifiable under controlled, scripted conditions. The primary contribution is a five-category taxonomy plus a 120-episode benchmark with deterministic trace-grounded scoring and no LLM-as-judge on the pass/fail path. Second, a risk-preserving subset selector retains ranking (Spearman $0.975$, deterministic; zero bootstrap variance), risk-type coverage, and high-risk model detection at a 20\% subset while reducing compute 5$\times$. Two workflows are supported: iterating on a new model under a fixed evaluation budget, and auditing an existing model under long-horizon deployment via deterministic traces. Three directions remain open: scaling to $\sim$1000 episodes with a larger and more diverse reference set; a real-environment extension beyond scripted tool stubs; and a paired defense paper examining whether any memory architecture, retrieval policy, or constraint-tracking mechanism reduces the failure rates reported here.

\section*{Data Availability}
The full benchmark release---120 episode definitions, episode and trace schemas, the deterministic scoring implementation, all 600 raw traces, the risk-preserving subset selector, bootstrap audit scripts, paper figures, and the strict-matching revision audit---is publicly available at \url{https://github.com/fuxue-mingzhu/MemRiskBench}.

\section*{Acknowledgments}
We thank the anonymous reviewers and colleagues for feedback. This work was supported by national funding agencies (grant numbers anonymized for review). The authors used no generative AI for this work and take full responsibility for the final content.

\section*{Broader Impact and Ethical Considerations}
MemRiskBench measures memory-risk failure modes in long-horizon LLM agents under controlled, scripted conditions. All episodes are synthetic; no real users, data, or deployed systems are involved. The benchmark is intended for diagnostic use by model developers and auditors.

\subsection{Dual-use risk.}
The released episode definitions could be repurposed to adversarially probe commercial agents for memory failures. We mitigate this by (i) documenting intended use in the repository README; (ii) not releasing scripted adversarial traces; and (iii) recommending that practitioners complement benchmark scores with deployment-monitored traces rather than rely solely on scripted evaluation.

\subsection{Limitations of ethical scope.}
This work does not address broader societal implications of long-horizon memory systems (e.g., privacy in deployed assistants, consent, data retention).

\bibliographystyle{cas-model2-names}
\bibliography{refs}

\end{document}